\documentclass[letterpaper, 10 pt, conference]{URL-ieeeconf}
\IEEEoverridecommandlockouts    % This command is only needed if
\usepackage{graphics}           
\usepackage{times}              
\usepackage{amsmath}            
\usepackage{amssymb}            
\usepackage{graphicx}
\usepackage{algorithm}
\usepackage[noend]{algpseudocode}
\usepackage{booktabs}
\usepackage{color}
\usepackage{multirow}
\usepackage{subcaption}
\usepackage{rotating}
\usepackage{cite}
\definecolor{instructioncolor}{rgb}{.5,.5,.5}

\def\eqref#1{(\ref{#1})}

\newcommand{\rom}[1]{\uppercase\expandafter{\romannumeral #1\relax}}

\makeatletter
\usepackage{xspace}
\DeclareRobustCommand\onedot{\futurelet\@let@token\@onedot}
\def\@onedot{\ifx\@let@token.\else.\null\fi\xspace}

\makeatother

\usepackage{array}
\newcolumntype{L}[1]{>{\raggedright\let\newline\\\arraybackslash\hspace{0pt}}m{#1}}
\newcolumntype{C}[1]{>{\centering\let\newline\\\arraybackslash\hspace{0pt}}m{#1}}
\newcolumntype{R}[1]{>{\raggedleft\let\newline\\\arraybackslash\hspace{0pt}}m{#1}}

\newcommand{\inv}{^{-1}}

\newcommand{\pvnbin}{B} % Total number of bins
\newcommand{\pvnt}{(x, y)}
\newcommand{\pvnnt}{n\pvnt}

\newcommand{\pvpk}{p_k\pvnt}
\newcommand{\pvck}{c_k\pvnt}
\newcommand{\pvw}{W_{\mathrm{time}}}

\newcommand{\pvpkset}{\{\pvpk\}_{k=0}^{\pvnbin-1}}

\newcommand{\pvh}{h\pvnt}

\newcommand{\pvq}{q\pvnt}
\newcommand{\pvs}{s\pvnt}

\newcommand{\wk}{\omega_{k}}
\newcommand{\wmin}{\omega_{\mathrm{min}}}
\newcommand{\ww}{w}
\newcommand{\wpt}{w_{\mathrm{pt}}}
\newcommand{\wline}{w_{\mathrm{line}}}
\newcommand{\zz}{w_{\mathrm{EV}}}
\newcommand{\zpt}{w_{\mathrm{EV}_{\mathrm{pt}}}}
\newcommand{\zline}{w_{\mathrm{EV}_{\mathrm{line}}}}
\newcommand{\wba}{w_{\mathrm{BA}}}
\newcommand{\wbapt}{w_{\mathrm{BA}_{\mathrm{pt}}}}
\newcommand{\wbaline}{w_{\mathrm{BA}_{\mathrm{line}}}}
\newcommand{\alphapt}{\alpha_{\mathrm{pt}}}
\newcommand{\alphaline}{\alpha_{\mathrm{line}}}

\newcommand{\gdist}{\gamma_{\mathrm{dist}}}
\newcommand{\OmL}{\Omega_{\ell}}
\newcommand{\sline}{S_{\mathrm{line}}}

\newcommand{\fpt}{\mathcal{F}_{\mathrm{pt}}}
\newcommand{\fline}{\mathcal{F}_{\mathrm{line}}}
\newcommand{\rpt}{\mathbf{r}_\mathrm{pt}}
\newcommand{\rline}{\mathbf{r}_{\mathrm{line}}}
\newcommand{\rprior}{\mathbf{r}_{\mathrm{p}}}
\newcommand{\hprior}{\mathbf{H}_{\mathrm{p}}}
\newcommand{\rimu}{\mathbf{r}_{\mathrm{I}}}
\newcommand{\pimu}{\mathbf{P}_{\mathrm{I}}}

\newcommand{\Twb}{\mathbf{T}_{wb}}
\newcommand{\Tbc}{\mathbf{T}_{bc}}
\newcommand{\pw}{\mathbf{p}_{w}}

\newcommand{\us}{\mathbf{u}_{\mathrm{s}}}
\newcommand{\ue}{\mathbf{u}_{\mathrm{e}}}

\newcommand{\lprojb}{\mathbf{l}^{\prime} = [l_{1}^{\prime}, l_{2}^{\prime}, l_{3}^{\prime}]^\top}

\newcommand{\pn}{p_{n}}

\newcommand{\trackm}{\mathbf{m}_{\mathrm{track}}}
\newcommand{\trackmn}{\mathbf{m}_{n}}
\newcommand{\trackmt}{\mathbf{m}_{t}}

\newcommand{\avggrad}{\bar{\mathbf{g}}_{t}}

\newcommand{\wratio}{w_{\mathrm{ratio}}}
\newcommand{\wstatic}{w_{\mathrm{stat}}}
\newcommand{\wdynamic}{w_{\mathrm{dyn}}}
\makeatletter
\let\NAT@parse\undefined
\makeatother

\PassOptionsToPackage{colorlinks=true,citecolor=blue,linkcolor=blue, urlcolor=blue,bookmarks=false,hypertexnames=true}{hyperref}
\usepackage{etoolbox}
\usepackage{hyperref}

\apptocmd{\normalsize}{
  \setlength{\abovedisplayskip}{3pt plus 1pt minus 1pt}
  \setlength{\belowdisplayskip}{3pt plus 1pt minus 1pt}
  \setlength{\abovedisplayshortskip}{0pt plus 1pt}
  \setlength{\belowdisplayshortskip}{2pt plus 1pt minus 1pt}
}{}{}
\title{\LARGE \bf PLED-VINS: A Point-Line Event-Based Visual Inertial SLAM for Dynamic Environments}

\author{Seunghun Lee$^{1\dagger}$, Jihun Nam$^{2\dagger}$, Dong-Uk Seo$^{2}$, and Hyun Myung$^{2*}$, \IEEEmembership{Senior Member, IEEE}
\thanks{$^{*}$Corresponding author: Hyun Myung.}
\thanks{$^{\dagger}$Both authors have equally contributed.}%
\thanks{$^{1}$Robotics Program, KAIST (Korea Advanced Institute of Science and Technology), Daejeon, 34141, South Korea. {\tt\small shleee@kaist.ac.kr}}
\thanks{$^{2}$School of Electrical Engineering, KAIST (Korea Advanced Institute of Science and Technology), Daejeon, 34141, South Korea. {\tt\small \{namjh1228, dongukseo, hmyung\}@kaist.ac.kr}}
}

\begin{document}
\maketitle
\thispagestyle{empty}
\pagestyle{empty}

%%%%%%%%%%%%%%%%%%%%%%%%%%%%%%%%%%%%%%%%%%%%%%%%%%%%%%%%%%%%%%%%%%%%%%%%%%%%%%%%
\begin{abstract}
Dynamic environments remain a fundamental challenge for visual SLAM, where unreliable observations from moving objects and rapid motion degrade state estimation accuracy. Although event cameras preserve fine-grained spatio-temporal information, most existing event-based SLAM frameworks still assume static scenes and lack approaches to estimate the reliability of features.
To this end, we propose~\textit{PLED-VINS}, a monocular event camera-based visual-inertial SLAM framework that enables robust state estimation in dynamic environments. We propose an entropy-recency score map to characterize the temporal reliability of both point and line features based on event temporal statistics. Concurrently, geometric reliability is estimated via a unified point–line robust bundle adjustment. Building upon these, we design an adaptive weighting strategy that fuses temporal and geometric reliability, including motion-conditioned reliability modeling for line features, to suppress unreliable observations. 
Experimental results demonstrate that \textit{PLED-VINS} improves state estimation on the evaluated dynamic sequences with moving objects.
\end{abstract}

%%%%%%%%%%%%%%%%%%%%%%%%%%%%%%%%%%%%%%%%%%%%%%%%%%%%%%%%%%%%%%%%%%%%%%%%%%%%%%%%
\section{Introduction}
\label{sec:intro}

Most visual simultaneous localization and mapping (SLAM) systems have been developed under the assumption of a static environment. This assumption, however, is frequently violated in dynamic scenes, where moving objects introduce erroneous geometric constraints during optimization, degrading pose estimation accuracy. Although several frame-based dynamic SLAM methods have attempted to identify dynamic objects or adjust feature weights~\cite{bescos2018ral, yu2018iros, song2022ral, wang2022iros, yin2022tro, zhang2024tim, huang2025applied}, these methods remain fragile in highly aggressive scenarios, due to motion-blurred frame observations.

\begin{figure}[t!]
    \centering
    \includegraphics[width=1.0\columnwidth, trim=10 25 0 50, clip]{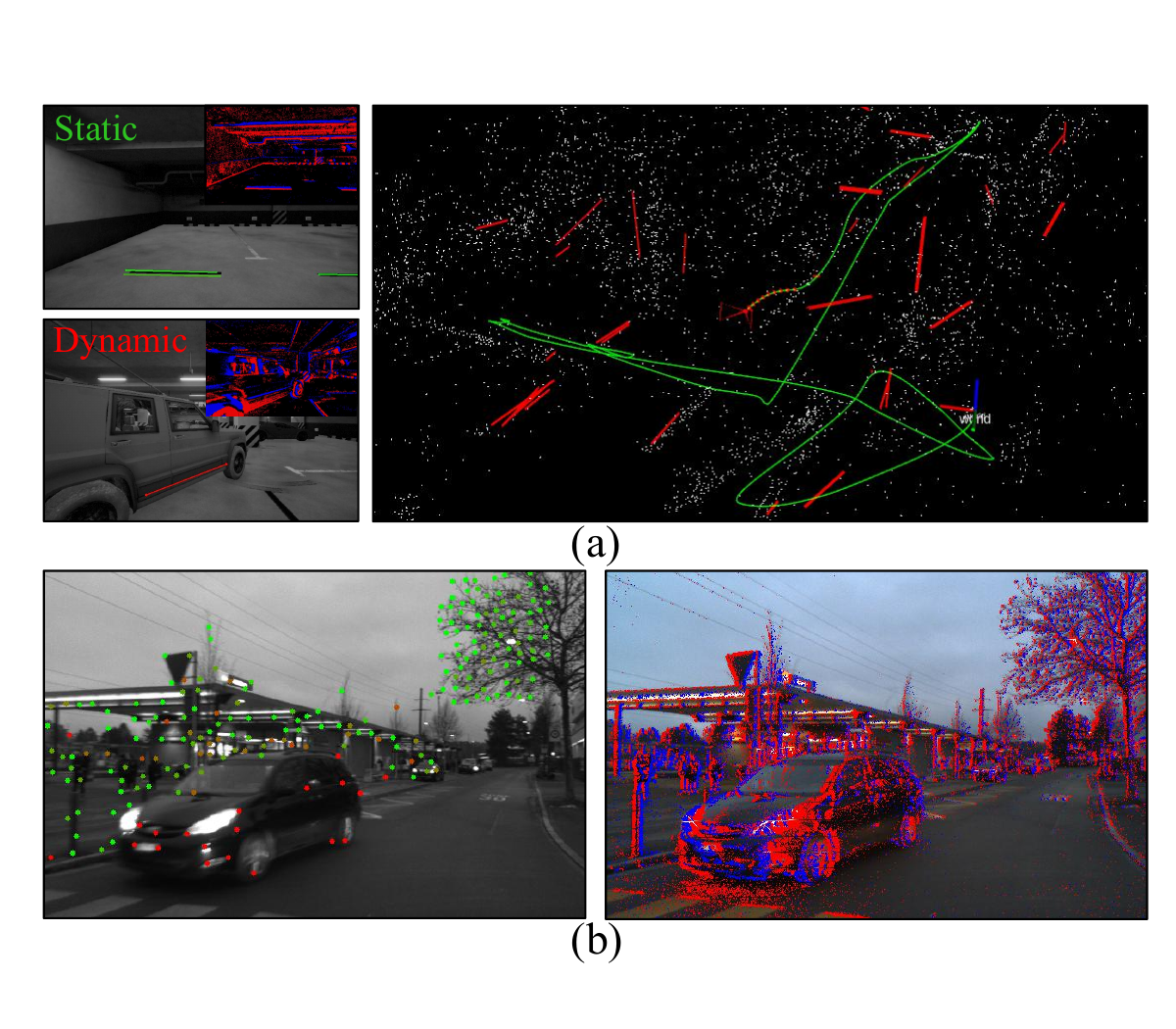}\vspace{-1.7mm}
    \caption{Our algorithm, \textit{PLED-VINS}, in dynamic environments. (a) Weighted line features and estimated trajectory on \texttt{parking\_lot high} sequence of VIODE dataset~\cite{minoda2021ral}. (b) Weighted point features and event streams on \texttt{zurich\_city\_01\_e} sequence of DSEC dataset~\cite{gehrig2021ral}. Green and red indicate high and low feature weights, respectively.}
    \label{fig:pledvins}
\end{figure}

\begin{figure*}[t!]
    \centering
    \includegraphics[width=1.0\textwidth, trim= 20 100 90 37, clip]{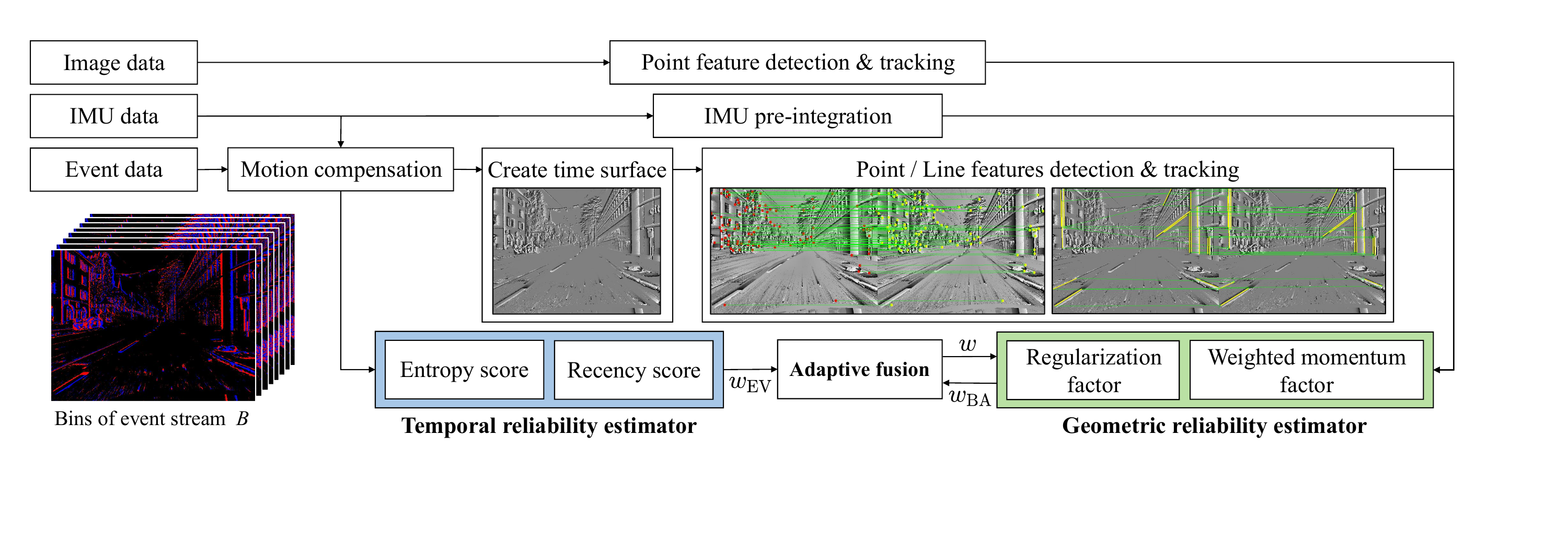}
    \caption{
    The architecture of the proposed PLED-VINS. The system takes monocular images, IMU, and raw event data as inputs. The temporal reliability estimator divides the event stream into temporal bins $B$ to compute an entropy–recency score map, producing weights $\zz$. Concurrently, the geometric reliability estimator evaluates geometric reliability via a line-augmented robust BA, producing weights $\wba$. These cues are adaptively fused to obtain weights $\ww=\{\wpt, \wline\}$, which are fed back into the subsequent BA for robust state estimation.}
    \label{fig:system_overview}
\end{figure*}

Event cameras~\cite{gallego2020survey} help address the limitations of frame-based cameras in low-light and high-speed motion scenes. By asynchronously recording per-pixel brightness changes, event cameras achieve high temporal resolution, enabling precise analysis of instantaneous motion. Despite the advantages, the majority of event-based SLAM frameworks~\cite{rebcq2017ral, vidal2018ral, guan2023tase, chen2023ral, tang2024iros} assume static environments. One promising direction is to integrate event-based motion segmentation into SLAM for dynamic scenes. Existing methods based on contrast maximization and clustering~\cite{stoffregen2019iccv, zhou2023tnnls, zhao2023icra}, however, are computationally expensive and introduce substantial integration overhead, limiting their real-time applicability for state estimation. Thus, a unified event-based SLAM framework that efficiently incorporates dynamic information remains underexplored.

In this context, we propose \textit{PLED-VINS}, a robust monocular event camera-based visual-inertial SLAM framework for dynamic environments. As illustrated in Fig.~\ref{fig:pledvins}, PLED-VINS incorporates feature-level temporal and bundle adjustment (BA)-based geometric reliability cues for real-time performance instead of explicit object-level segmentation. Event cameras offer rich temporal dynamics, yet their potential for dynamic SLAM remains underexplored. To leverage this capability, we construct an entropy–recency score map that captures the distribution and recency of events at each pixel, thereby quantifying feature-level temporal reliability. We further estimate geometric reliability as feature weights optimized in a unified point–line robust BA~\cite{song2022ral}, and adaptively fuse the temporal and geometric cues to weight point and line features, improving pose estimation robustness. Our contributions are summarized as follows:
\begin{itemize}
    \item We propose a novel entropy–recency score map that captures temporal distributions of event streams to estimate the reliability of dynamic observations.
    \item We develop an adaptive weighting strategy that combines temporal and geometric reliability to jointly reflect the consistency of observations.
    \item We evaluate PLED-VINS against existing methods on dynamic sequences, showing improved state estimation in scenes with moving objects.
\end{itemize}

%%%%%%%%%%%%%%%%%%%%%%%%%%%%%%%%%%%%%%%%%%%%%%%%%%%%%%%%%%%%%%%%%%%%%%%%%%%%%%%%
\section{Related Works}
\label{sec:related}
\subsection{Feature-based Event SLAM}
Feature-based event SLAM methods~\cite{vidal2018ral, chen2023ral, guan2023tase, tang2024iros} estimate camera poses by extracting and tracking event features. Event cameras asynchronously generate a stream of events $\mathcal{E} = \{e_i\}_{i=1}^N$. Each event~\mbox{$e_i = \{x_i, y_i, t_i, p_i\}$} consists of the pixel coordinates  $u_i = (x_i,y_i)^T$, the timestamp $t_i$, and the polarity $p_i \in \{+1, -1\}$ indicating the direction of brightness change. To adopt events for SLAM frameworks, a surface of active events (SAE)~\cite{benosman2013event} is typically maintained. The SAE stores the timestamp of the most recent event at the pixel location $\mathbf{u}$, denoted as $t_{\mathrm{last}}(\mathbf{u})$. Based on SAE, the time surface (TS)~\cite{lagorce2017hots} is widely used to encode the recency of events. The TS at time $t \ge t_{\mathrm{last}}(\mathbf{u})$ is defined as:
\begin{equation}
    TS(\mathbf{u}, t) = \exp\left( - \frac{t - t_{\mathrm{last}}(\mathbf{u})}{\eta} \right),
\end{equation}
where $\eta$ is the decay rate parameter.

Using these representations, point features~\cite{vasco2016fast, alzugaray2018ral} and line features~\cite{vongioi2010tpami} are commonly extracted from the TS or SAE, while direct raw-event approaches have also been explored~\cite{hu2022iros}. In practice, many event-based SLAM systems primarily rely on point features, whose scarcity in low-texture environments can degrade pose accuracy. To mitigate this, several studies~\cite{guan2023tase, choi2025ral} integrate point and line features for more reliable geometric constraints. Nevertheless, these frameworks assume static environments, leaving robust handling of dynamic scenes an open challenge.

\subsection{Robust SLAM in Dynamic Environments}
Most visual SLAM frameworks assume static environments, resulting in errors in dynamic scenes, whereas dynamic SLAM improves robustness by detecting dynamic objects via semantic or geometric methods. Semantic approaches~\cite{yu2018iros, bescos2018ral, wang2022iros} leverage deep learning-based segmentation to detect dynamic objects. Bescos \textit{et al.}~\cite{bescos2018ral} adopted Mask R-CNN with multi-view geometry, Yu \textit{et al.}~\cite{yu2018iros} utilized SegNet, and Wang \textit{et al.}~\cite{wang2022iros} integrated point, line, and plane features with semantic and multi-view constraints. However, a common limitation of these methods is that they can only extract segmentation information for predefined classes.

Geometric approaches~\cite{zhang2024tim, yin2022tro, song2022ral, song2024ral, fu2022vins_dimc} distinguish dynamic objects based on geometric constraints, offering real-time efficiency and generalization. Song \textit{et al.}~\cite{song2022ral, song2024ral} mitigated dynamic features via IMU-aided optimization. Yin \textit{et al.}~\cite{yin2022tro} utilized virtual landmarks from stereo scene flow and IMU data. Zhang \textit{et al.}~\cite{zhang2024tim} integrated point and line features with a dynamic grid algorithm. Fu \textit{et al.}~\cite{fu2022vins_dimc} refined feature matching through multiple constraints, including flow vector bound. However, these methods often struggle in aggressive scenes, where rapid motion induces severe motion blur that can lead to incorrect geometric constraints.

In dynamic scenes, point-based methods are prone to data sparsity~\cite{song2022ral, yin2022tro, fu2022vins_dimc}, as suppressing dynamic observations reduces valid static ones. To address this, several frame-based studies~\cite{wang2022iros, zhang2024tim} have explored line features as complementary structural cues, providing stronger geometric constraints when point features are insufficient. However, most line-based dynamic SLAM methods rely on stereo or RGB-D configurations for stable geometric consistency enabled by explicit depth. In monocular systems, reliably evaluating line-based motion consistency remains challenging due to scale ambiguity and motion degeneracy.

These limitations motivate moving beyond purely geometric approaches for modeling feature reliability. A recent method leverages event cameras for temporal modeling~\cite{huang2025applied}, but compresses temporal information into a single average value. Consequently, different temporal behaviors may become indistinguishable, limiting reliable dynamic object discrimination in complex scenes.

%%%%%%%%%%%%%%%%%%%%%%%%%%%%%%%%%%%%%%%%%%%%%%%%%%%%%%%%%%%%%%%%%%%%%%%%%%%%%%%%
\section{PLED-VINS}
\label{sec:main}

\subsection{System Overview}\label{subsec:overview} 
The overall pipeline is illustrated in Fig.~\ref{fig:system_overview}, and our framework is based on DynaVINS~\cite{song2022ral}. Given incoming event and IMU measurements, IMU-based motion compensation on raw events~\cite{zhao2023icra} is first performed. Following this, the compensated events are encoded into a time surface (TS)~\cite{lagorce2017hots} via a surface of active events (SAE)~\cite{benosman2013event}. In parallel, IMU measurements are pre-integrated~\cite{forster2016manifold} for the sliding-window VIO backend.
For image data, only point features are extracted using the Shi--Tomasi~\cite{shi1994good} and tracked with Kanade--Lucas--Tomasi (KLT)~\cite{lucas1981iterative}. For event data, point and line features are extracted from TS using Arc*~\cite{alzugaray2018ral} and the line segment detector (LSD)~\cite{vongioi2010tpami}, respectively, with lines tracked by the line band descriptor~\cite{zhang2013jvcir}. To handle dynamic features, we construct an entropy-recency score map from compensated events (Section~\ref{subsec:entropymap}) based on their ego-motion consistency and use it to estimate the temporal reliability of point and line features (Section~\ref{subsec:temporalrel}).
In the backend (Section~\ref{subsec:geometricrel}), the sliding-window VIO jointly optimizes visual-inertial measurements while estimating geometric reliability by extending robust BA in DynaVINS~\cite{song2022ral} with weighted line factors. The temporal and geometric reliabilities are then adaptively fused through recursive refinement (Section~\ref{subsec:fusionopt}) to determine feature weights, yielding stable trajectory estimation in dynamic scenes.

\begin{figure}[t!]
    \centering
    \includegraphics[width=1.0\columnwidth, trim=40 146 40 147, clip]{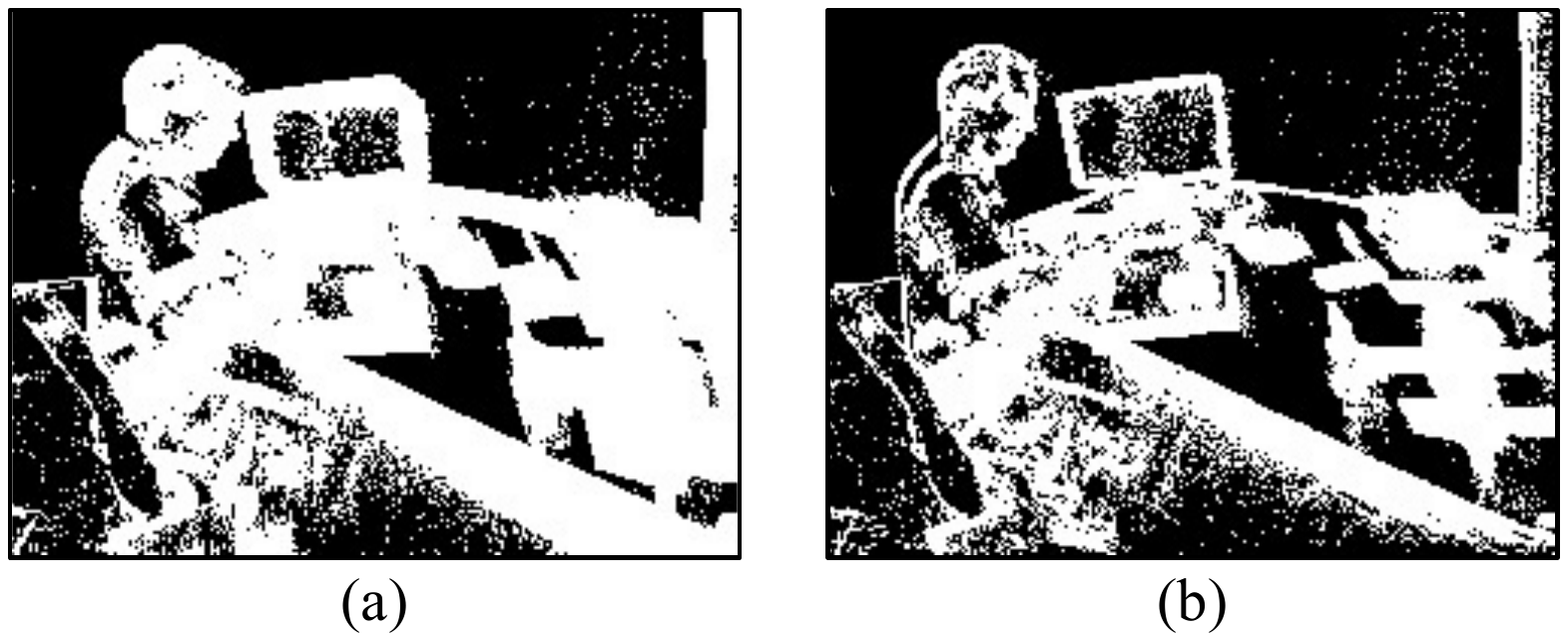}\vspace{-2mm}
    \caption{Comparison of IMU-based motion compensation. (a) Raw event streams. (b) Compensated event streams with sharpen edges for pixel-wise temporal analysis.}
    \label{fig:compensate}
\end{figure}

\subsection{Entropy-Recency Score Map}\label{subsec:entropymap}
In dynamic environments, independently moving objects often produce temporal behaviors that cannot be fully explained by geometric motion consistency alone. To capture such pixel-wise temporal characteristics in asynchronous event streams, we introduce the~\textit{entropy--recency score map}. This representation integrates information-theoretic entropy~\cite{shannon1948mathmatical} with temporal recency to characterize the stability of motion-compensated event distributions.

\textbf{Motion compensation:} 
Event streams are generated by both independently moving objects and the camera's ego-motion. In the presence of rotation, even static scene structures produce events that are spatially displaced over time. If such ego-motion is not compensated, these events spread across different pixels, distorting temporal statistics. To restore spatial alignment, we warp events to a reference frame using IMU measurements~\cite{zhao2023icra}. As shown in Fig.~\ref{fig:compensate}, this compensation sharpens event structures that are consistent with the dominant camera motion.

\textbf{Entropy score:} 
To estimate feature-level temporal reliability, we analyze the temporal distribution of events after motion compensation. For static scene structures, motion compensation aligns events at consistent pixels, producing a structured timestamp distribution consistent with a single rigid camera motion hypothesis. In contrast, events from independently moving objects remain inconsistent with the compensated motion, resulting in an unstructured temporal distribution even after alignment.

Specifically, for the motion-compensated event streams, the time window $\pvw$ is uniformly divided into $B$ consecutive bins. The probability mass function for each pixel is defined as 
$\pvpk = \pvck / (\pvnnt + \epsilon)$ for $k \in \{0, \dots, B-1\}$, 
where $\pvck$ is the event count in the $k$-th time bin, 
$\pvnnt$ is the total event count within the time window $\pvw$, 
and $\epsilon$ is a small positive constant. The resulting distribution $\pvpkset$ represents how events are distributed along the temporal axis within the time window. To quantify temporal consistency at each pixel, we compute the normalized Shannon entropy~\cite{cover1999elements}:
\begin{equation}\label{entropy_score}
    \pvh = -\frac{1}{\log B} \sum_{k=0}^{B-1} \pvpk \log(\pvpk + \epsilon).
\end{equation}
A higher $\pvh$ indicates that the temporal distribution is more dispersed within the time window, suggesting weaker temporal concentration under the compensated motion.

\begin{figure}[t!]
    \centering
    \includegraphics[width=1.0\columnwidth, trim= 30 22 20 48, clip]{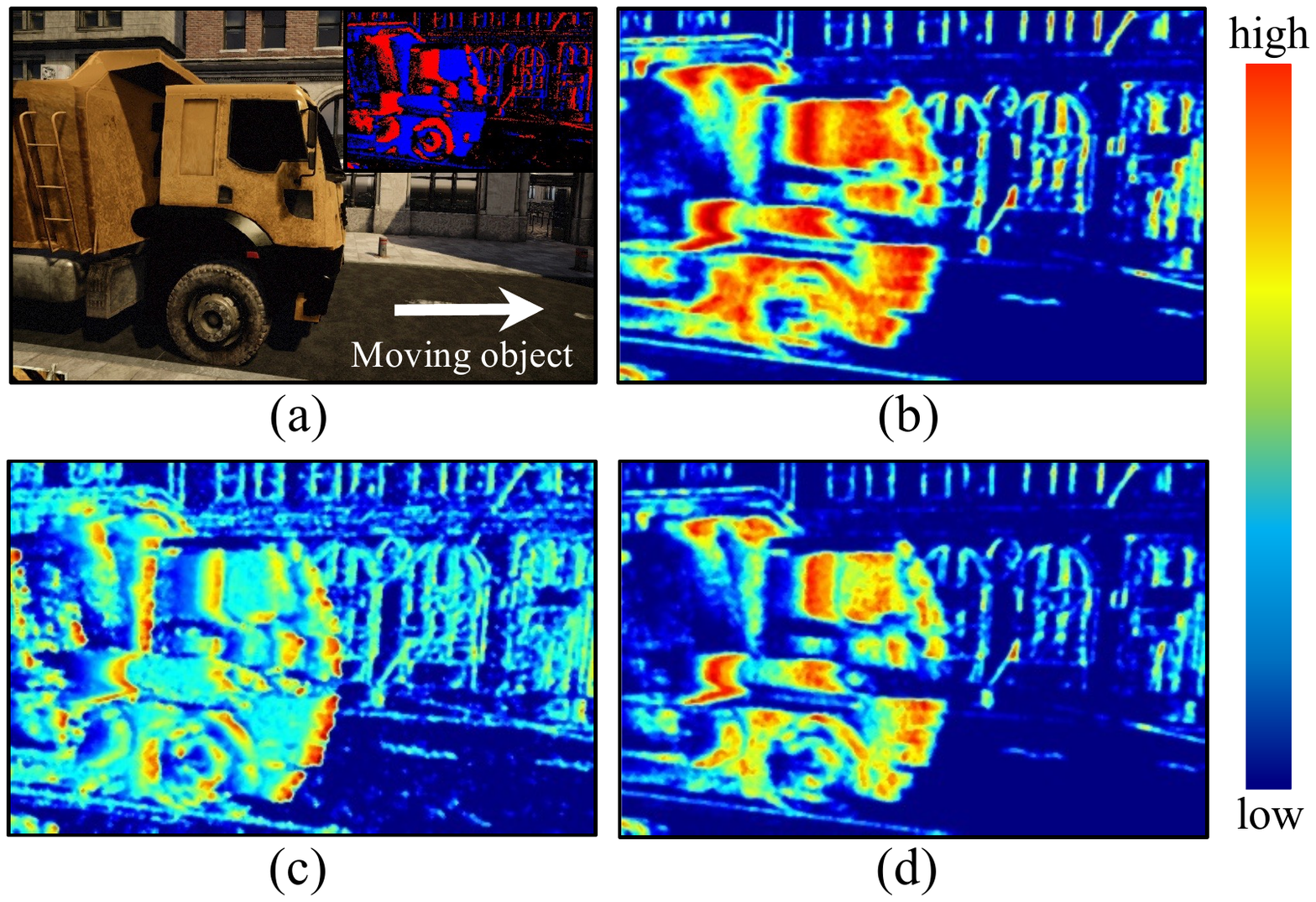}\vspace{-1mm}
    \caption{(a) Raw image and event. (b) Map of entropy score $\pvh$. (c) Map of recency score $\pvq$. (d) Map of entropy-recency score $\pvs$.}
    \label{fig:entropymap}
\end{figure}

However, as shown in Fig.~\hyperref[fig:entropymap]{\ref*{fig:entropymap}(b)}, elevated entropy responses may also arise in low-light or highly textured static background regions (e.g., the building edges on the right), where dense or noise-induced event generation spreads across multiple time bins. By incorporating the recency term described below, we suppress such temporally persistent activations.

\textbf{Recency score:} 
While entropy score in~\eqref{entropy_score} captures temporal dispersion, it ignores the temporal order of events. For static structures aligned by compensation, events accumulated at a fixed pixel are distributed over the time window rather than concentrated in the latest bins. In contrast, independently moving objects tend to produce late-biased activations after compensation. Prior work leverages the average timestamp as an indicator of dynamic regions~\cite{zhao2023icra}. However, such global averaging may be sensitive to the degree of dynamic activity and does not preserve the underlying temporal distribution. To explicitly model structured late activation without collapsing the distribution into a single statistic, we define the temporal recency score $\pvq$ as:
\begin{equation}
    \pvq = \frac{\sum_{k=0}^{B-1} \wk \pvpk - \wmin}{1 - \wmin + \epsilon}, \label{eq:recency}
\end{equation}
with
\begin{equation}
    \wk = \exp\left(-\alpha \left(1 - \frac{k+0.5}{B}\right)\right), \label{eq:weight}
\end{equation}
where $\wmin = e^{-\alpha}$ is the minimum weight, $\alpha > 0$ is the temporal decay rate, and $\epsilon$ is a small positive constant. This formulation assigns larger values when the event distribution is biased toward later bins, capturing recent activation patterns that are less typical in temporally persistent static structures. As illustrated in Fig.~\hyperref[fig:entropymap]{\ref*{fig:entropymap}(c)}, high recency scores are concentrated along the edge structures of the moving object. %, where features are typically extracted.

\textbf{Entropy-recency score:} 
Entropy and recency capture complementary aspects of temporal behavior. Entropy reflects deviations from motion-consistent temporal structure, while recency measures bias toward recent activation. Pixels with jointly high entropy and recency exhibit weak consistency with the single-motion assumption and a bias toward recent activation, characteristics commonly associated with independently moving objects. To jointly reflect these complementary cues into a unified entropy–recency score, we employ the following nonlinear symmetric fusion:
\begin{equation}
    \pvs = \frac{\pvh \pvq}{\pvh \pvq + (1 - \pvh)(1 - \pvq)}.
    \label{eq:final_fusion_score}
\end{equation}
The fusion in~\eqref{eq:final_fusion_score} emphasizes strong agreement between entropy and recency while suppressing cases where only one indicator is elevated, yielding a normalized score for temporal inconsistency. Unlike a simple product, the normalization promotes symmetric agreement and suppresses unilateral elevation. As shown in Fig.~\hyperref[fig:entropymap]{\ref*{fig:entropymap}(d)}, the fused score suppresses static background regions and low-texture areas while clearly highlighting the independently moving object.

%%%%%%%%%%%%%%%%%%%%%%%%%%%%%%%%%%%%%%%%%%%%%%%%%%%%%%%%%%%%%%%%%%%%%%%%%%%%%%%%%%%%%%%%%%%%%%%%%%%%%%%%%%%%%%%%%%%%%

\subsection{Temporal Reliability Estimation}\label{subsec:temporalrel}

% We compute the temporal reliability of both point and line features from the entropy–recency score map, leveraging temporal information to reduce the influence of dynamic observations.

We compute the temporal reliability of point and line features using the entropy–recency score map to suppress dynamic observations.

\textbf{Point feature: }The temporal reliabilities of the points are evaluated using~\eqref{eq:final_fusion_score}. For a point feature located at coordinates $\pvnt$, the temporal reliability is defined as \mbox{$\zpt\pvnt = 1 - \pvs$}, representing the probability that the point belongs to a static structure. 

\textbf{Line feature: } Unlike point features, line measurements have spatial extent. Therefore, temporal reliability is computed by aggregating entropy–recency statistics within a narrow neighborhood around each line, denoted as the line band region $\OmL$.

Concretely, given a line feature $\ell = (\mathbf{p}_0, \mathbf{p}_1)$ extracted from the TS, where $\mathbf{p}_0, \mathbf{p}_1 \in \mathbb{R}^2$ denote its endpoints, 
$\OmL$ is defined as a band region parameterized by the unit direction vector $\mathbf{v}$ and the orthogonal vector $\mathbf{n}$:
\begin{equation}
\OmL \! = \! \{ \mathbf{x} \!= \mathbf{p}_0 \! + \! \lambda L \mathbf{v} \! + \! \delta\mathbf{n} \mid 
\lambda \in [0,1],\;
\delta \in \left[-\tfrac{W}{2}, \tfrac{W}{2}\right] \}
\end{equation}
where $\lambda$ is the axial interpolation factor, $\delta$ is the orthogonal offset, $L = \|\mathbf{p}_1 - \mathbf{p}_0\|$ is the line length, $W$ is the band width, $\mathbf{v} = (\mathbf{p}_1 - \mathbf{p}_0)/L$ is the unit direction vector, and $\mathbf{n}$ is the unit normal vector orthogonal to $\mathbf{v}$.

To compute the entropy–recency score, $\sline$, for a line feature, we consider the set of pixels within $\OmL$, and utilize Gaussian-weighted averaging. A Gaussian distance weight $\gdist(\delta)$ based on the orthogonal offset $\delta$ is applied to emphasize samples closer to the line axis while suppressing peripheral noise. Accordingly, $\sline$ is defined as:
\begin{equation} 
\sline = \frac{\sum_{\mathbf{x} \in \OmL \cap \mathbb{Z}^2} \gdist(\delta(\mathbf{x})) \cdot s(\mathbf{x})}{\sum_{\mathbf{x} \in \OmL \cap \mathbb{Z}^2} \gdist(\delta(\mathbf{x}))}, 
\end{equation}
where $s(\mathbf{x})$ is the entropy–recency score of pixel $\mathbf{x}$, and \mbox{$\OmL \cap \mathbb{Z}^2$} represents the set of image pixels within $\OmL$. The line temporal reliability is defined as $\zline = 1 - \sline$.

The temporal reliabilities $\zpt$ and $\zline$ are jointly integrated with the geometric reliability in \mbox{Section~\ref{subsec:geometricrel}} to determine the point and line feature weights \mbox{$\wpt, \wline \in [0,1]$} through the adaptive fusion in \mbox{Section~\ref{subsec:fusionopt}}.

%%%%%%%%%%%%%%%%%%%%%%%%%%%%%%%%%%%%%%%%%%%%%%%%%%%%%%%%%%%%%%%%%%%%%%%%%%%%%%%%%%%%%

\subsection{Geometric Reliability Estimation}\label{subsec:geometricrel}
Beyond temporal reliability, we model geometric reliability through a unified robust optimization for point and line features by extending~\cite{song2022ral} with line factors.
Specifically, we jointly optimize the sliding-window state $\mathcal{X}$, following~\cite{fu2020plvins}, together with the feature weights $\wpt$ and $\wline$ as:
\begin{equation}\label{eq:opt_expression}
\begin{aligned}
\min_{\mathcal{X}, \mathcal{W}} \Biggl\{ 
& \left\| \rprior - \hprior \mathcal{X} \right\|^2 
+ \sum_{k \in \mathcal{B}} \left\| \rimu^k(\mathcal{X}) \right\|_{\pimu}^2 \\[-6pt]
& + \sum_{j \in \fpt} \rho(\wpt^j, \rpt^j) 
+ \sum_{j \in \fline} \rho(\wline^j, \rline^j) \Biggr\},
\end{aligned}
\end{equation}
where $\rprior$ and $\hprior$ represent the residual and measurement matrix of the marginalization prior, respectively. $\mathcal{B}$ is the set of IMU preintegration indices, and $\rimu^k$ is the IMU preintegration residual with covariance $\pimu$. The robust loss $\rho$ for feature $j$ is:
\begin{equation}
\begin{aligned}
\rho(w^j, \mathbf{r}^j) = & (w^j)^2 \sum_{i \in \mathcal{P}(f_j)} | \mathbf{r}^{j,i}(\mathcal{X}) |^2 \\
& + \lambda_w (1-w^j)^2 + \lambda_m (n_j(\tilde{w}^j-w^j))^2,
\end{aligned}
\end{equation}
where $w^j$ is the current weight and $\tilde{w}^j$ is its previous estimate. $\mathbf{r}^{j,i}$ denotes the visual residual in the $i$-th frame, $\mathcal{P}(f_j)$ is the set of observing frames, $n_j$ is the number of observations of the $j$-th feature, and
the hyperparameters $\lambda_w$ and $\lambda_m$ control their strengths~\cite{song2022ral}.

For a point feature $j$ observed in the $i$-th frame, the reprojection residual $\rpt^{j,i}$ is defined as the difference between the observed coordinate $\mathbf{u}_j^i$ and its reprojected position:
\begin{equation}
\rpt^{j,i} = \mathbf{u}_j^i - \pi \left( \Tbc\inv (\Twb^{i})\inv \pw^j \right),
\end{equation}
where $\pi(\cdot)$ is the camera projection model, $(\Twb^{i})\inv$ is the body pose at the $i$-th frame, $\Tbc$ is the fixed extrinsic, and $\pw^j$ is the 3D position of feature $j$ in the world frame.

For a line feature $l$ observed in the $i$-th frame, the reprojection residual $\rline^{l,i}$~\cite{fu2020plvins} is the perpendicular distance from endpoints $\us^{l,i}, \ue^{l,i}$ to the projected line \mbox{$\lprojb$}:
\begin{equation}
\label{eq:line_residual}
\rline^{l,i} = \left[ 
\frac{(\us^{l,i})^\top \mathbf{l}^{\prime}}{\sqrt{{l}_{1}^{\prime 2} + {l}_{2}^{\prime 2}}}, \quad 
\frac{(\ue^{l,i})^\top \mathbf{l}^{\prime}}{\sqrt{{l}_{1}^{\prime 2} + {l}_{2}^{\prime 2}}} 
\right]^\top.
\end{equation}
% The optimized weights $\wbapt$ and $\wbaline$ serve as geometric reliability measures in the temporal–geometric fusion described in the following section.
Subsequently, the optimized weights $\wpt$ and $\wline$ in~\eqref{eq:opt_expression}  serve as geometric reliability measures in Section~\ref{subsec:fusionopt}.
%%%%%%%%%%%%%%%%%%%%%%%%%%%%%%%%%%%%%%%%%%%%%%%%%%%%%%%%%%%%%%%%%%%%%%%%%%%%%%%%
\begin{figure}[t!]
    \centering
    \includegraphics[width=1.0\columnwidth, trim=90 20 239 0, clip]{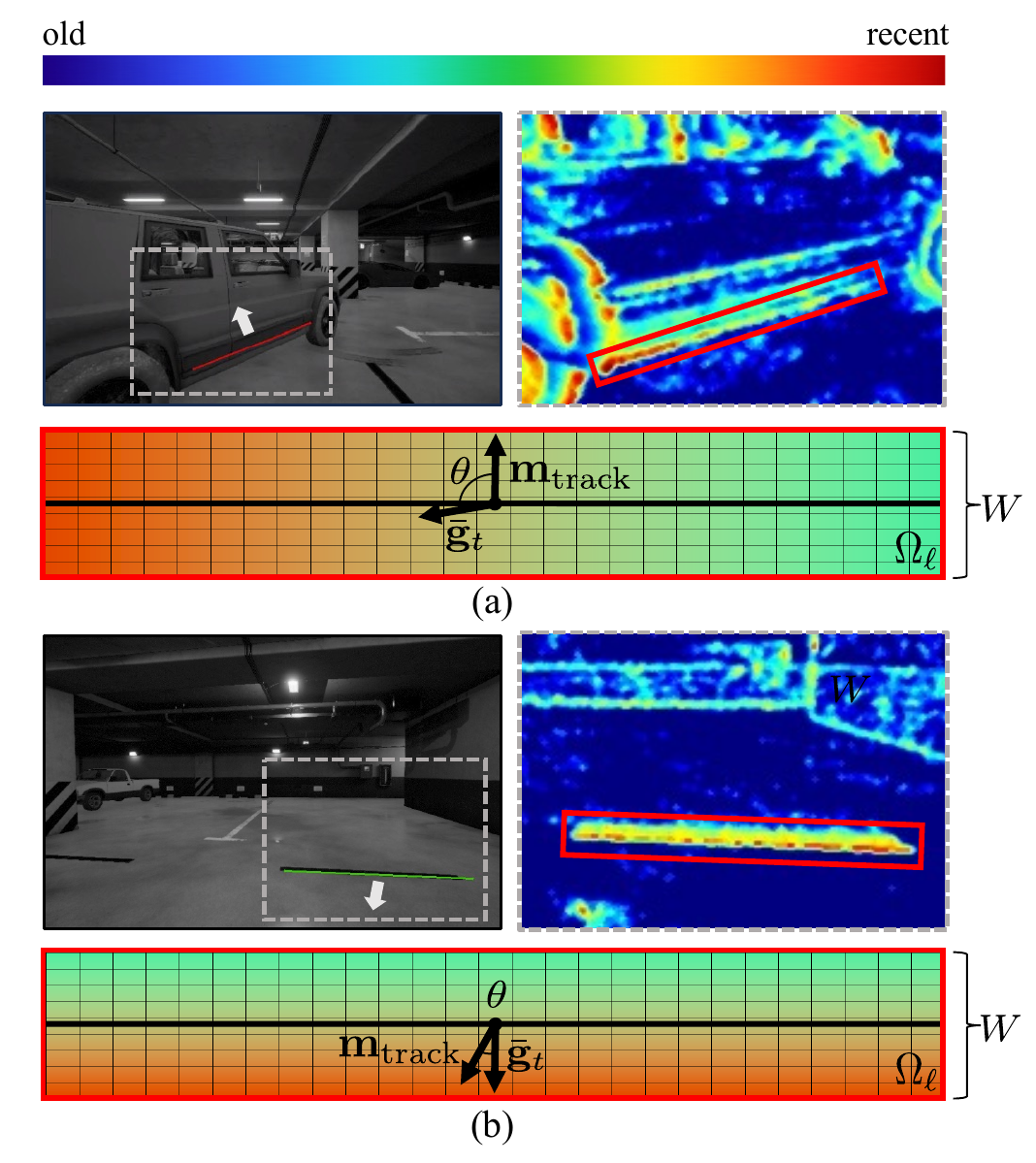}\vspace{-3.5mm}
    \caption{Gradient alignment within the line band region. 
    Mean temporal gradient $\avggrad$ on the recency score map $\pvq$ and tracking motion direction $\trackm$. 
    (a) Motion and gradient are misaligned, inconsistent with the assumed motion hypothesis. 
    (b) Motion aligned with the temporal gradient structure, indicating motion-consistent temporal behavior.}
    \label{fig:line}
\end{figure}
%%%%%%%%%%%%%%%%%%%%%%%%%%%%%%%%%%%%%%%%%%%%%%%%%%%%%%%%%%%%%%%%%%%%%%%%%%%%%%%%

\subsection{Adaptive Fusion with Temporal and Geometric Reliability}\label{subsec:fusionopt}
Temporal reliability in Section~\ref{subsec:temporalrel} captures motion-consistent timing in event streams, while geometric reliability in Section~\ref{subsec:geometricrel} reflects reprojection consistency in BA. Since they assess observation consistency from different domains, their joint consideration enables a more comprehensive reliability estimation. To this end, we introduce an adaptive fusion scheme to refine feature weights. The point and line weights $\wpt$ and $\wline$ are computed as:
\begin{equation}\label{eq:new_wpt}
    \wpt = (1 - \alphapt) \wbapt + \alphapt \zpt,
\end{equation}
\begin{equation}\label{eq:new_wline}
    \wline = (1 - \alphaline) \wbaline + \alphaline \zline,
\end{equation}
where $\alphapt$ and $\alphaline$ are adaptive coefficients controlling the contribution of temporal reliability. 
The geometric reliability cues $\wbapt$ and $\wbaline$ correspond to the optimized weights $\wpt$ and $\wline$ for points and lines in~\eqref{eq:opt_expression}, respectively. Subsequently, the fused weights from~\eqref{eq:new_wpt} and~\eqref{eq:new_wline} are used in the next BA iteration of~\eqref{eq:opt_expression}, enabling recurrent refinement.

\textbf{Fusion coefficient for point features: }
Using geometric reliability $\wbapt$ and temporal reliability $\zpt$, the fusion coefficient $\alphapt$ is defined as:
\begin{equation}
\alphapt = \alpha_0 \exp\!\left(-\frac{|\zpt - \wbapt|}{\tau}\right),
\end{equation}
where $\alpha_0$ is the base fusion rate and $\tau$ controls the decay. This formulation reduces $\alphapt$ when $\zpt$ and $\wbapt$ differ significantly, limiting the contribution of temporal reliability and mitigating the effect of temporal-geometric disagreement on the fused weight.

\textbf{Fusion coefficient for line features: } 
 The reliability of line features depends on the interaction between camera motion and line geometry. Since the line residual in~\eqref{eq:line_residual} only accounts for the perpendicular distance, motion normal to the line provides stronger geometric constraints, whereas tangential motion yields weak observability. 

 To capture this, we define a motion-conditioned fusion coefficient $\alphaline$ that integrates geometric observability and temporal consistency, defined as:
\begin{equation}
\alphaline = 1 - \pn (1 - r),
\end{equation}
where $\pn$ denotes the normal flow ratio and $r$ represents the gradient alignment term.
By design, $\alphaline$ increases when $r$ indicates strong alignment and remains high under tangential motion, where the normal flow ratio $\pn$ is small.

To reflect geometric observability, we define the normal flow ratio $\pn$ as the proportion of the tracking motion projected onto the line normal direction.
Let $\trackm$ denote the 2D tracking motion direction of the line estimated from feature tracking, decomposed into normal and tangential components as $\trackm = \trackmn + \trackmt$. The normal flow ratio is then defined as:
\begin{equation}
\pn = \frac{\left\| \trackmn \right\|}{\left\| \trackmn \right\| + \left\| \trackmt \right\| + \epsilon},
\end{equation}
which represents the relative magnitude of the normal motion component. As defined in~\eqref{eq:line_residual}, the line projection residual measures the perpendicular distance to the projected line; thus, increasing the normal motion component enhances residual sensitivity and geometric discriminability. Under such conditions, the line residual becomes more informative, so the fused weight in~\eqref{eq:new_wline} places greater emphasis on the geometric reliability term $\wbaline$ by reducing $\alphaline$.

We then define the gradient alignment term $r$ to quantify whether temporal evidence supports the estimated motion as:
\begin{equation}
r = \frac{|\avggrad \cdot \trackm|}{\|\avggrad\| \|\trackm\| + \epsilon}, 
\quad 
\avggrad = \frac{1}{|\OmL|} \sum_{\mathbf{x} \in \OmL} \nabla q(\mathbf{x}),
\end{equation}
where $\avggrad$ is the mean temporal gradient within the line band region $\OmL$. 
A higher $r$ indicates stronger agreement between the temporal gradient structure and the tracking motion direction. 
Under a motion-consistent hypothesis, the temporal gradient aligns with the dominant motion direction.
As illustrated in Fig.~\ref{fig:line}, alignment reflects motion-consistent temporal behavior, whereas misalignment indicates inconsistency with the assumed motion model, in which case the fusion shifts greater reliance to geometric evidence. Consequently, $\alphaline$ increases when temporal evidence aligns with the motion pattern and decreases when geometric observability is strong, as reliable geometric constraints diminish the reliance on temporal cues. It thus adaptively modulates the weights of temporal and geometric reliability in~\eqref{eq:new_wline}, shifting $\wline$ between $\zline$ and $\wbaline$ depending on motion conditions.

%%%%%%%%%%%%%%%%%%%%%%%%%%%%%%%%%%%%%%%%%%%%%%%%%%%%%%%%%%%%%%%%%%%%%%%%%%%%%%%%
\begin{table*}[t!]
    \centering
    \captionsetup{font=footnotesize}
    \caption{Absolute trajectory error (ATE, m) on the VIODE dataset~\cite{minoda2021ral} across different dynamic levels, with $\lambda_w = 2.0$ and $\lambda_m = 0.2$. The setup column denotes sensor types (E: Event, V: Visual, I: Inertial), camera configurations (M: Monocular, S: Stereo), and utilized features (P: Point, L: Line). The best results are highlighted in \textbf{bold}, while the second-best are \underline{underlined}.}\vspace{-1.5mm}
    \label{table:viode_full_results}
    \resizebox{\textwidth}{!}{
        \begin{tabular}{l p{1.0cm} p{1.4cm} cccccccccccc}
            \toprule \midrule
            \multirow{2}{*}[-0.5ex]{Method} &
            \multicolumn{2}{c}{Setup} &
            \multicolumn{4}{c}{city\_day} &
            \multicolumn{4}{c}{city\_night} &
            \multicolumn{4}{c}{parking\_lot} \\
            \cmidrule(lr){2-3} \cmidrule(lr){4-7} \cmidrule(lr){8-11} \cmidrule(lr){12-15}
            & \raggedright Sensor & \raggedright Cam / Feat
            & \texttt{none} & \texttt{low} & \texttt{mid} & \texttt{high}
            & \texttt{none} & \texttt{low} & \texttt{mid} & \texttt{high}
            & \texttt{none} & \texttt{low} & \texttt{mid} & \texttt{high} \\
            \midrule
            PL-VINS~\cite{fu2020plvins}  & V-I    & M / P+L & 0.435 & 0.219 & 0.741 & 7.102 & 0.709 & 0.732 & 1.441 & 2.393 & 0.151 & --- & --- & --- \\
            DynaVINS~\cite{song2022ral} & V-I    & M / P   & \underline{0.201} & \textbf{0.181} & \underline{0.189} & 0.226 & 0.339 & \underline{0.302} & \underline{0.425} & \underline{0.284} & \underline{0.093} & 0.216 & 0.175 & \underline{0.200} \\
            \midrule
            PL-EVIO~\cite{guan2023tase}  & E-V-I & M / P+L & 0.322 & 0.666 & 1.219 & 3.215 & 1.255 & 1.568 & 1.796 & 1.451 & 1.066 & 1.017 & 5.343 & 2.576 \\
            E2-VINS~\cite{huang2025applied}  & E-V-I & M / P   & 0.280 & 0.328 & 0.475 & \underline{0.225} & 0.712 & 1.014 & 1.306 & 0.566 & 0.396 & \textbf{0.123} & 0.606 & 0.422 \\
            ESVIO~\cite{chen2023ral}    & E-V-I & S / P   & 0.432 & 0.451 & 0.585 & 0.559 & \underline{0.299} & 0.767 & 1.048 & 1.418 & 0.170 & \underline{0.143} & \underline{0.162} & 0.237 \\
            \midrule
            \textbf{Ours} & E-V-I & M / P+L
            & \textbf{0.184} & \underline{0.215} & \textbf{0.157} & \textbf{0.170}
            & \textbf{0.290} & \textbf{0.262} & \textbf{0.125} & \textbf{0.137}
            & \textbf{0.076} & 0.183 & \textbf{0.087} & \textbf{0.076} \\
            \midrule
            \bottomrule
        \end{tabular}
    }
\end{table*}

\begin{figure}[t!]\vspace{-1mm}
    \centering
    \includegraphics[width=1.0\columnwidth, trim = 0 7 5 7, clip]{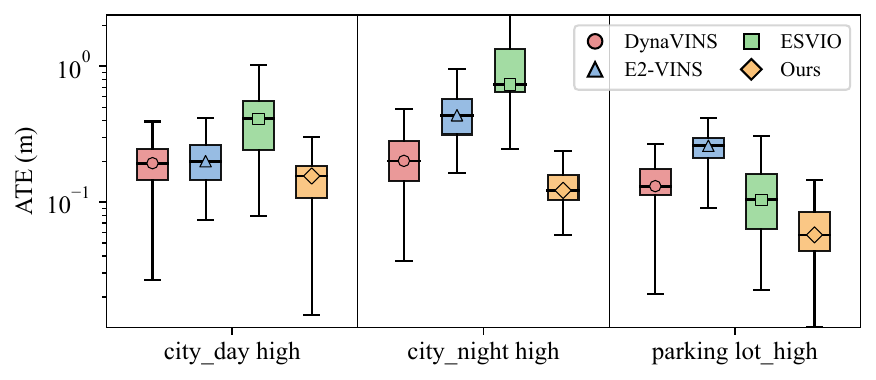}\vspace{-1.5mm}
    \caption{ATE comparison on the  \texttt{high} sequences of VIODE dataset~\cite{minoda2021ral}.}
    \label{fig:ATE_results}
\end{figure}

\section{Experimental Evaluation}
\label{sec:exp}

\subsection{Experimental Settings and Metrics}\label{subsec:exp_settings}
All experiments were conducted on a desktop PC with an AMD Ryzen 9 5950X CPU. For the proposed method, we used a TS decay of $\eta=20\,\mathrm{ms}$, $B=8$ temporal bins, $\alpha=2.0$ for recency weighting, and $\alpha_0=0.4$, $\tau=0.2$ for adaptive fusion. Trajectory accuracy was evaluated using absolute trajectory error (ATE, m) and mean position error (MPE, \%/m), with all trajectories aligned to ground truth via SE(3) Umeyama alignment~\cite{umeyama1991tpami}. We compared ours against representative feature-based open-source baselines: \mbox{PL-VINS}~\cite{fu2020plvins}, \mbox{DynaVINS}~\cite{song2022ral}, \mbox{PL-EVIO}~\cite{guan2023tase}, \mbox{E2-VINS}~\cite{huang2025applied}, \mbox{ESVIO}~\cite{chen2023ral}, and \mbox{Ultimate-SLAM}~\cite{vidal2018ral}.

\subsection{Datasets}\label{subsec:datasets}
\textbf{VIODE dataset:}
VIODE dataset~\cite{minoda2021ral} provides simulated image-IMU sequences with moving vehicles and dynamic occlusions. As it does not provide event data, synthetic events were generated using v2e~\cite{hu2021cvpr}. The subsequences ($\texttt{none}, \texttt{low}, \texttt{mid},$ and $\texttt{high}$) represent increasing dynamic levels, providing structured geometry for evaluating line feature integration.

\textbf{DAVIS 240C dataset:}
DAVIS 240C dataset~\cite{mueggler2017ijrr} provides real-world data with asynchronous events, images, IMU data, and ground-truth poses. The \texttt{dynamic\_translation} and \texttt{dynamic\_6dof} sequences involving a moving human subject were used for evaluation. These sequences include aggressive motion, making them suitable for evaluating robustness under highly dynamic conditions.

\textbf{DSEC dataset:}
DSEC dataset~\cite{gehrig2021ral} provides high-resolution driving sequences with image, event, IMU, and LiDAR data. As ground-truth poses are unavailable, qualitative evaluation was performed to assess feature reliability and visualize the estimated trajectories. Specifically, we employed highly dynamic sequences, \texttt{zurich\_city\_01\_e} and \texttt{zurich\_city\_01\_f}, to evaluate the stability of the proposed framework in outdoor scenarios while discriminating dynamic objects.

\begin{figure}[t!]\vspace{-1mm}
    \centering
    \includegraphics[width=1.0\columnwidth, trim = 20 10 50 39, clip]{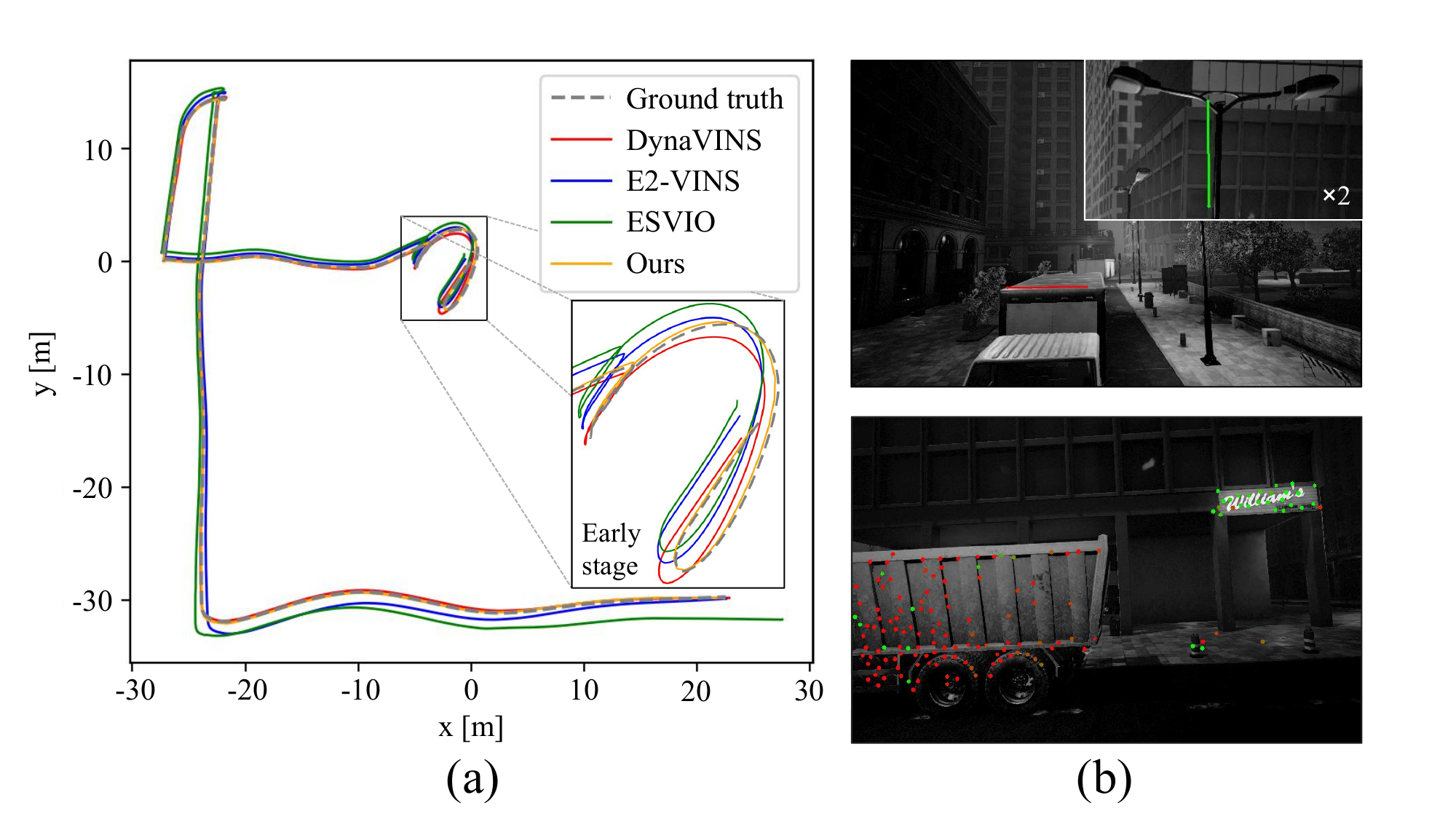}\vspace{-2mm}
    \caption{(a) Trajectory comparison on the \texttt{city\_night high} sequence of VIODE dataset~\cite{minoda2021ral}. (b) Feature weights in our proposed method (green: high, red: low).}
    \label{fig:traj_results}
\end{figure}

\subsection{Evaluation on the VIODE Dataset}\label{subsec:eval_viode} 
In the VIODE dataset, the number of dynamic objects increases across four levels (\texttt{none} to \texttt{high}). Table~\ref{table:viode_full_results} reports the RMSE of ATE across all dynamic levels. 
\mbox{Ultimate-SLAM~\cite{vidal2018ral}}, a monocular method based on event-image point features, was excluded from the comparison, as it failed to produce stable trajectory estimates on all sequences in the VIODE dataset.
Overall, the proposed framework delivered consistently improved accuracy across most sequences and dynamic levels. DynaVINS~\cite{song2022ral} suppresses dynamic observations via motion agreement but does not consider temporal information, resulting in consistently higher errors than our temporally-aware method. PL-VINS~\cite{fu2020plvins}, PL-EVIO~\cite{guan2023tase}, and ESVIO~\cite{chen2023ral} exhibited performance degradation in the \texttt{high} sequences. These degradations stem from their reliance on static-scene geometric assumptions. Although E2-VINS~\cite{huang2025applied} incorporates temporal cues to improve robustness under dynamic environments, it still degraded in these sequences, due to limited exploitation of temporal information.
In contrast, our method leveraged temporal reliability and achieved the lowest ATE in all \texttt{high} sequences as shown in Fig.~\ref{fig:ATE_results}.
 
Fig.~\ref{fig:traj_results} presents trajectory comparisons and corresponding point and line weights on the \texttt{city\_night high} sequence. The compared methods deviate from the ground truth along the early curved segment. In this stage, incomplete convergence of IMU biases and initialization errors can distort reprojection residuals. Estimation driven mainly by geometric consistency may therefore suffer unstable convergence under dynamic interference. By incorporating complementary temporal reliability, our method mitigates early residual distortion and stabilizes convergence under strong dynamics, remaining closely aligned with the ground truth.

% %%%%%%%%%%%%%%%%%%%%%%%%%%%%%%%%%%%%%%%%%%%%%%%%%%

\subsection{Evaluation on the DAVIS 240C Dataset}\label{subsec:eval_davis}
The evaluation on the DAVIS 240C dataset assesses trajectory estimation in the presence of dynamic objects and real-world sensor noise.
The \texttt{dynamic\_translation} and \texttt{dynamic\_6dof} sequences were evaluated using MPE normalized by traveled distance, providing a stable evaluation metric for the short trajectories.
As shown in Table~\ref{table:davis_results}, our method achieved the lowest MPE in both sequences, even in more challenging sequence \texttt{dynamic\_6dof}, outperforming competing methods. 
PL-VINS~\cite{fu2020plvins}, Ultimate-SLAM~\cite{vidal2018ral}, and PL-EVIO~\cite{guan2023tase} exhibit rapid error accumulation, while E2-VINS~\cite{huang2025applied} and DynaVINS~\cite{song2022ral} maintain partial robustness yet degrade under sustained dynamic conditions. In contrast, our method compensates for unstable geometric observations under aggressive motion by adaptively weighting features with complementary temporal reliability.

Qualitative results on the \texttt{dynamic\_translation} sequence in Fig.~\ref{fig:feat_results} show effective suppression of dynamic features while preserving static structures. Under this DAVIS 240C setting, the proposed method showed an average processing frequency of 21.25\,Hz at $240 \times 180$ resolution.

{\setlength{\floatsep}{4pt plus 1pt minus 1pt}
\begin{table}[t!]
    \centering
    \captionsetup{font=footnotesize}
    \caption{Mean position error (MPE, \%/m) on the DAVIS 240C dataset~\cite{mueggler2017ijrr}, with $\lambda_w = 8.0$ and $\lambda_m = 1.0$. The best results are highlighted in \textbf{bold}, while the second-best are \underline{underlined}.}\vspace{-1.5mm}
    \label{table:davis_results}
    \resizebox{\columnwidth}{!}{
        \begin{tabular}{lcc}
            \toprule \midrule
            Method & \texttt{dynamic\_translation} & \texttt{dynamic\_6dof} \\ \midrule
            PL-VINS~\cite{fu2020plvins}        & 0.158 & 0.402 \\
            Ultimate-SLAM~\cite{vidal2018ral}  & 0.263 & 0.360 \\
            PL-EVIO~\cite{guan2023tase}        & 0.175 & 0.653 \\
            E2-VINS~\cite{huang2025applied}        & \underline{0.121} & 0.292 \\
            DynaVINS~\cite{song2022ral}       & 0.139 & \underline{0.217} \\ \midrule
            \textbf{Ours}  & \textbf{0.089} & \textbf{0.176} \\ \midrule
            \bottomrule
        \end{tabular}
    }

\end{table}

\begin{figure}[t!]\vspace{-2.5mm}
    \centering
    \includegraphics[width=1.0\columnwidth, trim = 0 27 25 10, clip]{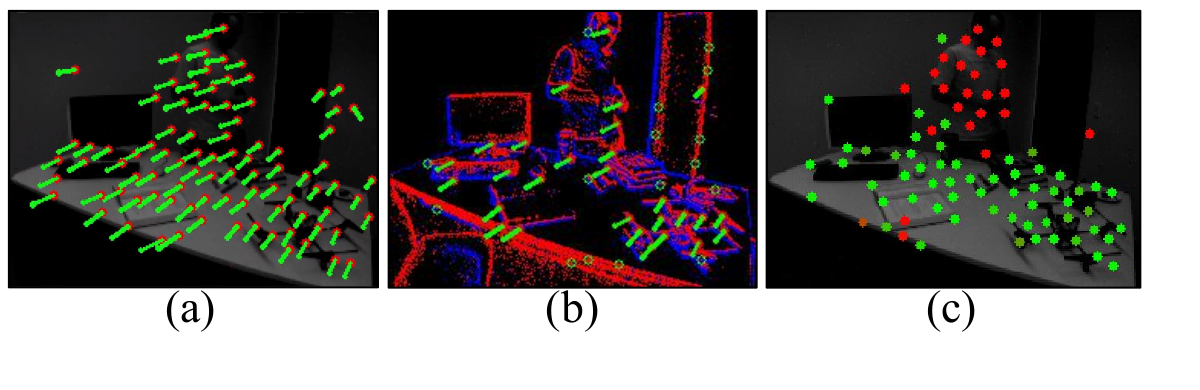}\vspace{-1mm}
    \caption{Feature detection and weighting on the DAVIS 240C dataset. 
    (a) Image-based detection. 
    (b) Event-based detection. 
    (c) Representation of feature weights (green: high, red: low).}
    \label{fig:feat_results}
\end{figure}
}
\vspace{0.5mm}
\subsection{Evaluation on the DSEC Dataset}\label{subsec:eval_dsec}
\begin{figure}[t!]
    % \vspace{-3mm}
    \centering
    \includegraphics[width=1.0\columnwidth, trim = 0 227 0 36, clip]{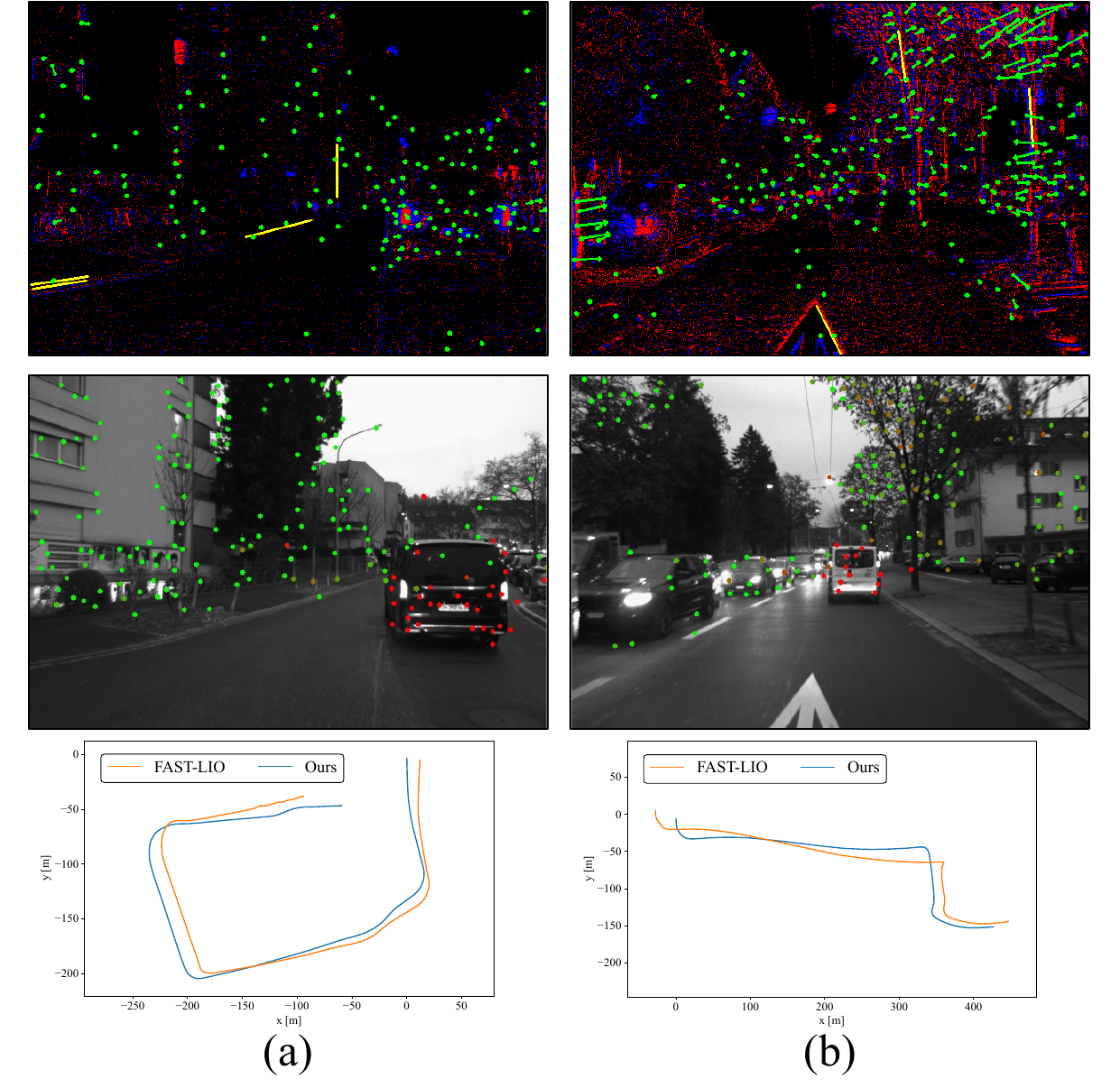}\vspace{-2mm}
    \caption{
    Qualitative results of PLED-VINS on the DSEC dataset~\cite{gehrig2021ral}:
    (a) \texttt{zurich\_city\_01\_e}.
    (b) \texttt{zurich\_city\_01\_f}.
    \textbf{Top}: Point and line feature tracking results on the event stream.
    \textbf{Middle}: Feature weights (green: high, red: low)
    \textbf{Bottom}: Estimated trajectories of the proposed method and FAST-LIO~\cite{xu2021ral}.
    }
    \label{fig:dsec_results}
\end{figure}

Fig.~\ref{fig:dsec_results} illustrates qualitative results of our framework on the DSEC dataset~\cite{gehrig2021ral}, with \mbox{$\lambda_w = 2.0$} and \mbox{$\lambda_m = 2.0$}. These results qualitatively validate the proposed reliability estimator across driving scenarios with dynamic objects. In \texttt{zurich\_city\_01\_e} and \texttt{zurich\_city\_01\_f}, our framework effectively suppressed features associated with moving vehicles, retaining reliable features from the static background. Ground-truth poses are not provided in DSEC dataset; thus, we did not perform quantitative comparisons against ground-truth. Instead, the trajectories were compared with a FAST-LIO reference trajectory~\cite{xu2021ral}. 
The results show plausible motion patterns consistent with the driving scenarios and close agreement between the two methods, despite the highly dynamic nature of the evaluated sequences.

\begin{table}[t!]
\centering
\captionsetup{font=footnotesize}
\caption{Ablation study on \texttt{parking\_lot high} of VIODE dataset~\cite{minoda2021ral}. 
Weight quality is evaluated using MAE ($\downarrow$) and weight ratio $w_{\mathrm{ratio}}$ ($\uparrow$).}\vspace{-1.5mm}
\label{table:ablation_dynamic}
\resizebox{\columnwidth}{!}{
\begin{tabular}{l cc cc}
\toprule \midrule
\multirow{2}{*}[-0.6ex]{Method}
& \multicolumn{2}{c}{Point} 
& \multicolumn{2}{c}{Line} \\
\cmidrule(lr){2-3} \cmidrule(lr){4-5}
& MAE $\downarrow$ & $w_{\mathrm{ratio}}$ $\uparrow$  
& MAE $\downarrow$ & $w_{\mathrm{ratio}}$ $\uparrow$\\
\midrule
Geometric-only 
& 0.449 & 1.577 
& 0.627 & 1.162 \\

Geometric + Temporal 
& \textbf{0.382} & \textbf{1.870} 
& \textbf{0.339} & \textbf{2.151} \\
\midrule
\bottomrule
\end{tabular}
}
\end{table}
\vspace{0.5mm}
\subsection{Ablation Study}\label{subsec:ablation_studies}
To analyze the contribution of the temporal reliability estimator, we compared two configurations: geometric-only weighting based solely on robust BA reliability, and the full model integrating temporal and geometric reliability. To assess weighting quality without fixed thresholds, we used two metrics. First, the mean absolute error (MAE) between estimated weights and ground-truth dynamic object labels from VIODE segmentation annotations~\cite{minoda2021ral} measures how well the continuous weights align with the ground-truth static/dynamic labels, where static labels correspond to a target weight of 1 and dynamic labels to 0. Second, we define the weight ratio \mbox{$\wratio = \mathrm{mean}(\wstatic) / \mathrm{mean}(\wdynamic)$}, where $\wstatic$ and $\wdynamic$ are the estimated weights of ground-truth static and dynamic features, respectively. A higher $\wratio$ indicates more effective suppression of dynamic observations relative to static ones.

As shown in Table~\ref{table:ablation_dynamic}, integrating temporal reliability reduced MAE and increased $\wratio$ for both point and line features. The improvement was particularly pronounced for line features, as incorporating temporal information more effectively suppressed dynamic observations insufficiently distinguished by geometric reliability alone.

%%%%%%%%%%%%%%%%%%%%%%%%%%%%%%%%%%%%%%%%%%%%%%%%%%%%%%%%%%%%%%%%%%%%%%%%%%%%%%%%
\section{Conclusion}
\label{sec:conclusion}

In this paper, we presented PLED-VINS, a monocular event-based visual–inertial SLAM framework that models feature reliability in dynamic environments. By introducing an entropy–recency score map, we quantify temporal reliability from motion-compensated events and fuse them with geometric reliability, including motion-conditioned weighting for line features. This unified formulation suppresses dynamic observations while preserving stable structures. Experimental results demonstrate consistent improvements under highly dynamic scenarios. However, our framework assumes a dominant rigid-motion hypothesis for temporal modeling, which may be less reliable under depth-dependent parallax effects or complex multi-body dynamics. Future work will extend the reliability formulation to multi-motion hypotheses for clearer differentiation between ego-motion-consistent structures and independently moving objects.
%%%%%%%%%%%%%%%%%%%%%%%%%%%%%%%%%%%%%%%%%%%%%%%%%%%%%%%%%%%%%%%%%%%%%%%%%%%%%%%%
% Only if applicable
%\section*{Acknowledgments}
%We thank XXX for fruitful discussions and for \dots

\bibliographystyle{URL-IEEEtrans}

% All new citations should go to new.bib. The file glorified.bib should go
% be the one from the ipb server. After paper or related work has been
% written merge the entries from new.bib to glorified.bib ON THE SERVER,
% replace the glorified.bib in this repository and empty the new.bib
\bibliography{URL-bib}

\begin{thebibliography}{10}
\providecommand{\url}[1]{#1}
\csname url@rmstyle\endcsname
\providecommand{\newblock}{\relax}
\providecommand{\bibinfo}[2]{#2}
\providecommand\BIBentrySTDinterwordspacing{\spaceskip=0pt\relax}
\providecommand\BIBentryALTinterwordstretchfactor{4}
\providecommand\BIBentryALTinterwordspacing{\spaceskip=\fontdimen2\font plus
\BIBentryALTinterwordstretchfactor\fontdimen3\font minus \fontdimen4\font\relax}
\providecommand\BIBforeignlanguage[2]{{%
\expandafter\ifx\csname l@#1\endcsname\relax
\typeout{** WARNING: IEEEtran.bst: No hyphenation pattern has been}%
\typeout{** loaded for the language `#1'. Using the pattern for}%
\typeout{** the default language instead.}%
\else
\language=\csname l@#1\endcsname
\fi
#2}}

\bibitem{bescos2018ral}
B.~Bescos, J.~M. F{\'a}cil, J.~Civera, and J.~Neira, ``{DynaSLAM: Tracking, mapping, and inpainting in dynamic scenes},'' \emph{IEEE Robot. Automat. Lett.}, vol.~3, no.~4, pp. 4076--4083, 2018.

\bibitem{yu2018iros}
C.~Yu, Z.~Liu, X.~J. Liu, F.~Xie, Y.~Yang, Q.~Wei, and Q.~Fei, ``{DS-SLAM: A semantic visual SLAM towards dynamic environments},'' in \emph{Proc. IEEE/RSJ Int. Conf. Intell. Robot. Syst.}, 2018, pp. 1168--1174.

\bibitem{song2022ral}
S.~Song, H.~Lim, A.~J. Lee, and H.~Myung, ``{DynaVINS: A visual-inertial SLAM for dynamic environments},'' \emph{IEEE Robot. Automat. Lett.}, vol.~7, no.~4, pp. 11\,523--11\,530, 2022.

\bibitem{wang2022iros}
Y.~Wang, K.~Xu, Y.~Tian, and X.~Ding, ``{DRG-SLAM: A semantic RGB-D SLAM using geometric features for indoor dynamic scene},'' in \emph{Proc. IEEE/RSJ Int. Conf. Intell. Robot. Syst.}, 2022, pp. 1352--1359.

\bibitem{yin2022tro}
H.~Yin, S.~Li, Y.~Tao, J.~Guo, and B.~Huang, ``{Dynam-SLAM: An accurate, robust stereo visual-inertial SLAM method in dynamic environments},'' \emph{IEEE Trans. Robot.}, vol.~39, no.~1, pp. 289--308, 2022.

\bibitem{zhang2024tim}
B.~Zhang, X.~Ma, H.~J. Ma, and C.~Luo, ``{DynPL-SVO: A robust stereo visual odometry for dynamic scenes},'' \emph{IEEE Trans. Instrum. Meas.}, vol.~73, pp. 1--10, 2024.

\bibitem{huang2025applied}
J.~Huang, S.~Zhao, and L.~Zhang, ``{E2-VINS: An event-enhanced visual-inertial SLAM scheme for dynamic environments},'' \emph{Appl. Sci.}, vol.~15, no.~3, p. 1314, 2025.

\bibitem{minoda2021ral}
K.~Minoda, F.~Schilling, V.~W{\"u}est, D.~Floreano, and T.~Yairi, ``{VIODE: A simulated dataset to address the challenges of visual-inertial odometry in dynamic environments},'' \emph{IEEE Robot. Automat. Lett.}, vol.~6, no.~2, pp. 1343--1350, 2021.

\bibitem{gehrig2021ral}
M.~Gehrig, W.~Aarents, D.~Gehrig, and D.~Scaramuzza, ``{DSEC: A stereo event camera dataset for driving scenarios},'' \emph{IEEE Robot. Automat. Lett.}, vol.~6, no.~3, pp. 4947--4954, 2021.

\bibitem{gallego2020survey}
G.~Gallego, T.~Delbr{\"u}ck, G.~Orchard, C.~Bartolozzi, B.~Taba, A.~Censi, S.~Leutenegger, A.~J. Davison, J.~Conradt, K.~Daniilidis, and D.~Scaramuzza, ``{Event-based vision: A survey},'' \emph{IEEE Trans. Pattern Anal. Mach. Intell.}, vol.~44, no.~1, pp. 154--180, 2020.

\bibitem{rebcq2017ral}
H.~Rebecq, T.~Horstsch{\"a}fer, G.~Gallego, and D.~Scaramuzza, ``{EVO: A geometric approach to event-based 6-DOF parallel tracking and mapping in real time},'' \emph{IEEE Robot. Automat. Lett.}, vol.~2, no.~2, pp. 593--600, 2017.

\bibitem{vidal2018ral}
A.~R. Vidal, H.~Rebecq, T.~Horstsch{\"a}fer, and D.~Scaramuzza, ``{Ultimate SLAM? Combining Events, Images, and IMU for Robust Visual SLAM in HDR and High-Speed Scenarios},'' \emph{IEEE Robot. Automat. Lett.}, vol.~3, no.~2, pp. 994--1001, 2018.

\bibitem{guan2023tase}
W.~Guan, P.~Chen, Y.~Xie, and P.~Lu, ``{PL-EVIO: Robust monocular event-based visual inertial odometry with point and line features},'' \emph{IEEE Trans. Autom. Sci. Eng.}, vol.~21, no.~4, pp. 6277--6293, 2023.

\bibitem{chen2023ral}
P.~Chen, W.~Guan, and P.~Lu, ``{ESVIO: Event-based stereo visual inertial odometry},'' \emph{IEEE Robot. Automat. Lett.}, vol.~8, no.~6, pp. 3661--3668, 2023.

\bibitem{tang2024iros}
K.~Tang, X.~Lang, Y.~Ma, Y.~Huang, L.~Li, Y.~Liu, and J.~Lv, ``{Monocular event-inertial odometry with adaptive decay-based time surface and polarity-aware tracking},'' in \emph{Proc. IEEE/RSJ Int. Conf. Intell. Robot. Syst.}, 2024, pp. 12\,544--12\,551.

\bibitem{stoffregen2019iccv}
T.~Stoffregen, G.~Gallego, T.~Drummond, L.~Kleeman, and D.~Scaramuzza, ``{Event-based motion segmentation by motion compensation},'' in \emph{Proc. IEEE Int. Conf. Comput. Vis.}, 2019, pp. 7244--7253.

\bibitem{zhou2023tnnls}
Y.~Zhou, G.~Gallego, X.~Lu, S.~Liu, and S.~Shen, ``{Event-based motion segmentation with spatio-temporal graph cuts},'' \emph{IEEE Trans. Neural Netw. Learn. Syst.}, vol.~34, no.~8, pp. 4868--4880, 2023.

\bibitem{zhao2023icra}
C.~Zhao, Y.~Li, and Y.~Lyu, ``{Event-based real-time moving object detection based on IMU ego-motion compensation},'' in \emph{Proc. IEEE Int. Conf. Robot. Automat.}, 2023, pp. 690--696.

\bibitem{benosman2013event}
R.~Benosman, C.~Clercq, X.~Lagorce, S.-H. Ieng, and C.~Bartolozzi, ``{Event-based visual flow},'' \emph{IEEE Trans. Neural Netw. Learn. Syst.}, vol.~25, no.~2, pp. 407--417, 2013.

\bibitem{lagorce2017hots}
X.~Lagorce, G.~Orchard, F.~Galluppi, B.~E. Shi, and R.~B. Benosman, ``{HOTS: A hierarchy of event-based time-surfaces for pattern recognition},'' \emph{IEEE Trans. Pattern Anal. Mach. Intell.}, vol.~39, no.~7, pp. 1346--1359, 2017.

\bibitem{vasco2016fast}
V.~Vasco, A.~Glover, and C.~Bartolozzi, ``Fast event-based harris corner detection exploiting the advantages of event-driven cameras,'' in \emph{Proc. IEEE/RSJ Int. Conf. Intell. Robot. Syst.}, 2016, pp. 4144--4149.

\bibitem{alzugaray2018ral}
I.~Alzugaray and M.~Chli, ``{Asynchronous corner detection and tracking for event cameras in real-time},'' \emph{IEEE Robot. Automat. Lett.}, vol.~3, no.~4, pp. 3177--3184, 2018.

\bibitem{vongioi2010tpami}
R.~Grompone~von Gioi, J.~Jakubowicz, J.-M. Morel, and G.~Randall, ``{LSD: A fast line segment detector with a false detection control},'' \emph{IEEE Trans. Pattern Anal. Mach. Intell.}, vol.~32, no.~4, pp. 722--732, 2010.

\bibitem{hu2022iros}
S.~Hu, Y.~Kim, H.~Lim, A.~J. Lee, and H.~Myung, ``{eCDT: Event clustering for simultaneous feature detection and tracking},'' in \emph{Proc. IEEE/RSJ Int. Conf. Intell. Robot. Syst.}, 2022, pp. 3808--3815.

\bibitem{choi2025ral}
B.~Choi, H.~Lee, and C.~G. Park, ``{Event-frame-inertial odometry using point and line features based on coarse-to-fine motion compensation},'' \emph{IEEE Robot. Automat. Lett.}, vol.~10, no.~3, pp. 2622--2629, 2025.

\bibitem{song2024ral}
S.~Song, H.~Lim, A.~J. Lee, and H.~Myung, ``{DynaVINS++: Robust visual-inertial state estimator in dynamic environments by adaptive truncated least squares and stable state recovery},'' \emph{IEEE Robot. Automat. Lett.}, vol.~9, no.~10, pp. 9127--9134, 2024.

\bibitem{fu2022vins_dimc}
D.~Fu, H.~Xia, Y.~Liu, and Y.~Qiao, ``{VINS-Dimc: A visual-inertial navigation system for dynamic environment integrating multiple constraints},'' \emph{ISPRS Int. J. Geo-Inf.}, vol.~11, no.~2, p.~95, 2022.

\bibitem{forster2016manifold}
C.~Forster, L.~Carlone, F.~Dellaert, and D.~Scaramuzza, ``On-manifold preintegration for real-time visual--inertial odometry,'' \emph{IEEE Trans. Robot.}, vol.~33, no.~1, pp. 1--21, 2016.

\bibitem{shi1994good}
J.~Shi \emph{et~al.}, ``{Good features to track},'' in \emph{Proc. IEEE/CVF Conf. Comput. Vis. Pattern Recognit.}, 1994, pp. 593--600.

\bibitem{lucas1981iterative}
B.~D. Lucas and T.~Kanade, ``{An iterative image registration technique with an application to stereo vision},'' in \emph{Proc. Int. Joint Conf. on Artif. Intell.}, 1981, pp. 674--679.

\bibitem{zhang2013jvcir}
L.~Zhang and R.~Koch, ``{An efficient and robust line segment matching approach based on LBD descriptor and pairwise geometric consistency},'' \emph{J. Vis. Commun. Image Represent.}, vol.~24, no.~7, pp. 794--805, 2013.

\bibitem{shannon1948mathmatical}
C.~E. Shannon, ``A mathmatical theory of communication,'' \emph{The Bell System Technical Journal}, vol.~27, no.~3, pp. 370--423, 1948.

\bibitem{cover1999elements}
T.~M. Cover and J.~A. Thomas, \emph{Elements of Information Theory}.\hskip 1em plus 0.5em minus 0.4em\relax John Wiley \& Sons, 1999.

\bibitem{fu2020plvins}
Q.~Fu, J.~Wang, H.~Yu, I.~Ali, F.~Guo, Y.~He, and H.~Zhang, ``{PL-VINS: Real-time monocular visual-inertial SLAM with point and line features},'' \emph{arXiv preprint arXiv:2009.07462}, 2020.

\bibitem{umeyama1991tpami}
S.~Umeyama, ``{Least-squares estimation of transformation parameters between two point patterns},'' \emph{IEEE Trans. Pattern Anal. Mach. Intell.}, vol.~13, no.~4, pp. 376--380, 1991.

\bibitem{hu2021cvpr}
Y.~Hu, S.-C. Liu, and T.~Delbruck, ``{v2e: From video frames to realistic DVS events},'' in \emph{Proc. IEEE/CVF Conf. Comput. Vis. Pattern Recognit.}, 2021, pp. 1312--1321.

\bibitem{mueggler2017ijrr}
E.~Mueggler, H.~Rebecq, G.~Gallego, T.~Delbruck, and D.~Scaramuzza, ``{The event-camera dataset and simulator: Event-based data for pose estimation, visual odometry, and SLAM},'' \emph{Int. J. Robot. Res.}, vol.~36, no.~2, pp. 142--149, 2017.

\bibitem{xu2021ral}
W.~Xu and F.~Zhang, ``{Fast-LIO: A fast, robust LiDAR-inertial odometry package by tightly-coupled iterated Kalman filter},'' \emph{IEEE Robot. Automat. Lett.}, vol.~6, no.~2, pp. 3317--3324, 2021.

\end{thebibliography}

\end{document}